# Application of Machine Learning for Correcting Defect-induced Neuromorphic Circuit Inference Errors

Vedant Sawal, *Student Member, IEEE* and Hiu Yung Wong, *Senior Member, IEEE*

***Abstract*— This paper presents a machine learning-based approach to correct inference errors caused by stuck-at faults in fully analog ReRAM-based neuromorphic circuits. Using a Design-Technology Co-Optimization (DTCO) simulation framework, we model and analyze six spatial defect types—circular, circular-complement, ring, row, column, and checkerboard—across multiple layers of a multi-array neuromorphic architecture. We demonstrate that the proposed correction method, which employs a lightweight neural network trained on the circuit's output voltages, can recover up to 35% (from 55% to 90%) inference accuracy loss in defective scenarios. Our results, based on handwritten digit recognition tasks, show that even small corrective networks can significantly improve circuit robustness. This method offers a scalable and energy-efficient path toward enhanced yield and reliability for neuromorphic systems in edge and internet-of-things (IoTs) applications. In addition to correcting the specific defect types used during training, our method also demonstrates the ability to generalize—achieving reasonable accuracy when tested on *different types* of defects not seen during training. The framework can be readily extended to support real-time adaptive learning, enabling on-chip correction for dynamic or aging-induced fault profiles.**

## I. INTRODUCTION

NEUROMORPHIC computing promises to revolutionize edge and embedded intelligence by enabling energy-efficient inference at low latency [1]. Among emerging technologies, Resistive Random Access Memory (ReRAM) has shown strong potential for implementing analog compute-in-memory (CiM) architectures due to its non-volatility, scalability, and compatibility with CMOS processes [1][2]. In these systems, vector-matrix multiplication (VMM) operations are performed natively in the analog domain using Ohm's and Kirchhoff's laws, substantially reducing energy and area overhead compared to conventional digital accelerators.

Despite these advantages, the practical deployment of ReRAM-based neuromorphic circuits remains limited by their susceptibility to device-level variations and defects [3][4]. In particular, stuck-at faults resulting from fabrication imperfections or operational stress can significantly degrade inference accuracy [5][6]. Environmental/aging stress is also known to perturb ReRAM device characteristics and reliability [7][8][9]. Our studies have shown that defects such as stuck-off, stuck-on, or mismatched ReRAM pairs can introduce unpredictable voltage outputs that compromise prediction fidelity, especially in deeper neural layers [4]. Many existing studies or solutions often assume ideal or uniform defect distributions and fail to capture the structured, spatially correlated faults commonly observed in fabricated arrays [4][5][6].

In the literature, there have been works that investigated defect tolerance in ReRAM neuromorphic systems, including algorithm-level stuck-at fault mitigation [5], circuit and architecture co-design for high-defect scenarios [6], variation-aware readout circuits [3], and simulation frameworks analyzing inference robustness [10][11].

In our prior work, we characterized the degradation in inference accuracy caused by various spatially correlated faults (e.g., circular, ring, and column defects) in analog ReRAM arrays, and proposed a machine learning (ML) strategy to recover prediction accuracy based on the output node voltages of the faulty circuit [12]. However, that study was limited to isolated layers and fault types, and did not fully explore 'cross-defect generalization'—the ability of a corrective model trained on one type of defect (e.g., ring) to restore accuracy when tested on a different type (e.g., circle). Moreover, a correction NN which has a comparable size as or even larger than the neuromorphic network was used.

In this work, we present an extended and systematic investigation into the impact of stuck-at faults across all layers of a fully analog neuromorphic architecture. Using a Design-Technology Co-Optimization (DTCO) framework [10][13] with SPICE-level modeling of ReRAM devices, we analyze the role of defect type, size, and location. We also introduce a compact corrective neural network that learns to map faulty

Manuscript received on September 15, 2025. Vedant Sawal and Hiu Yung Wong are affiliated with the Multi-Physics And Circuit (M-PAC) Laboratory, Department of Electrical Engineering, San Jose State University, San Jose, CA 95192 USA. (Corresponding author: Hiu Yung Wong; e-mail: hiuyung.wong@sjsu.edu).

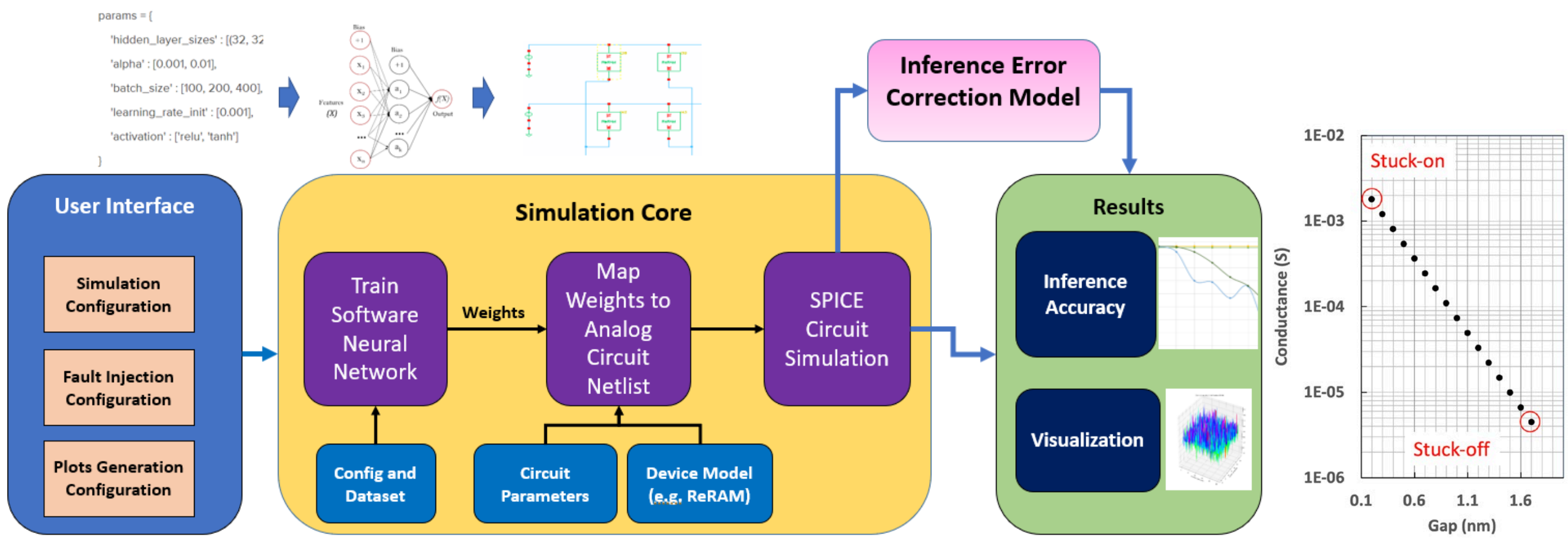

**Fig. 1:** Left: Design–Technology Co-Optimization (DTCO) framework. Right: ReRAM conductance vs. gap

outputs back to the correct class. The corrective network needs to be compact enough (many fewer neurons compared to the main circuit) to justify its usefulness. Our contributions are as follows:

1) We quantify the layer-wise sensitivity of inference accuracy to spatial defect type, size, and location, showing that both input- and output-layer faults can cause severe degradation.
2) We demonstrate that corrective models trained on one defect type can often generalize to other types and sizes, capturing robust fault-response patterns.
3) We show that even lightweight corrective networks (<200 parameters) can recover over 30% of lost accuracy in highly degraded cases, making them suitable for low-power edge deployment. This small corrective network can be fabricated using another technology (e.g., CMOS), which is more resilient to damage.

## II. Framework, Simulation Setup, Defect Modeling, and Data Generation

### *A. Simulation Framework and Setup*

To evaluate the impact of spatial defects in ReRAM-based neuromorphic circuits and propose correction strategies, we utilize a Design-Technology Co-Optimization (DTCO) simulation framework built with Python, interfaced with commercial SPICE simulators [12]. This end-to-end toolchain automates neural network training, circuit netlist generation, fault injection, simulation execution, and result visualization. An overview of the framework is shown in Fig. 1.

The simulation workflow begins with a software neural network (SNN) optimized for a given task using standard training techniques. The trained weights are then mapped to analog conductances that configure the ReRAM crossbar arrays in the neuromorphic circuit. Users define simulation parameters—including defect type, location, and magnitude—via structured JSON configuration files. The framework generates corresponding netlists that model fully analog circuit behavior, invokes SPICE-level simulation using Cadence Spectre [14], and extracts inference outputs for performance analysis.

For this study, we use the UCI handwritten digits dataset, consisting of 1797 grayscale images of size 8×8 pixels with 17 intensity levels. The neural network architecture comprises an input layer of 64 nodes, three hidden layers with 50, 20, and 8 neurons, and an output layer of 10 nodes—each corresponding to one digit (0–9). The total number of parameters in this network is 4528. The model is trained on 1617 images and tested on the remaining 180 images, achieving a baseline software inference accuracy of 96.67%. This network structure corresponds to four analog ReRAM arrays (layers 0 to 3) that implement the synaptic weights between adjacent layers via vector-matrix multiplication (VMM).

The neuromorphic circuit uses $HfO_X$-based ReRAM devices, modeled using a calibrated Verilog-A compact model that captures non-idealities such as gap size drift and temperature dependence [15]. The ReRAM device exhibits programmable conductance as a function of its internal gap size, with values ranging from 0.2 nm to 1.7 nm, as shown in Fig. 1. The analog arrays employ differential encoding for signed weight representation: each logical weight is implemented using the transconductances of a pair of physical ReRAMs, with their difference, $G_{ji}^{+} - G_{ji}^{-}$, representing the sign and magnitude of the weight at the logical row $j$ and logic column $i$. Each ReRAM array is followed by current comparators and rectifiers, enabling a fully analog signal path without digital conversion between layers.

In addition to the baseline inference path, the framework supports integration of a secondary neural network trained to correct inference errors based on the 10 output voltages of the neuromorphic circuit. This lightweight error-correction model is used to classify the correct digit even when faults disrupt the original prediction. In the framework, this part is implemented using software machine learning. In real deployment, it can be implemented by more robust technologies (e.g., CMOS neuromorphic circuit) or software ML as long as it is small enough, which will be shown to be possible.

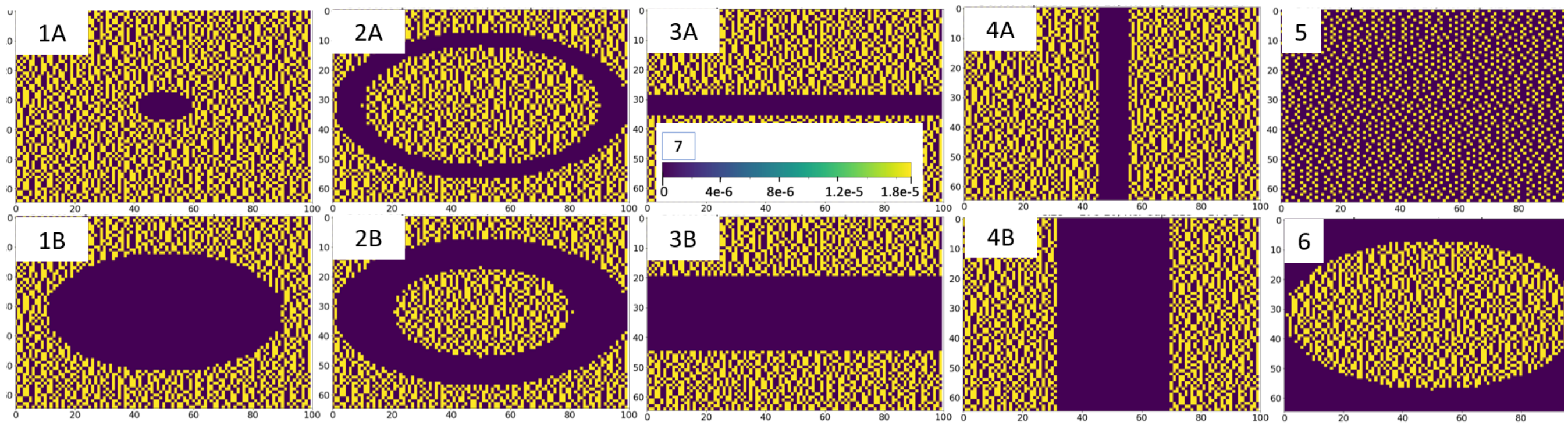

**Fig. 2:** Spatial defect patterns used in simulation, all shown for Layer 0 of the neuromorphic circuit. 1A–1B show circular defect masks with increasing radius $r \in \{0.1, 0.4\}$, expressed as a fraction of the array width, respectively. 2A–2B illustrate ring defects with an outer radius fixed at 0.5 and inner radii of 0.4 and 0.28. 3A–3B depict row defects with heights of 1 and 4 rows, while 4A–4B depict column defects with widths of 1 and 4 columns. 5 displays the checkerboard defect pattern, which features a fixed alternating pattern of stuck and non-stuck cells. 6 shows circle-complement defects where only the central circular region is functional. Color bar indicates conductance in Siemens (inset 7).

### *B. Defect Modeling*

In this study, a stuck-on fault is modeled as the gap size being stuck at 0.2nm (~1.8mS) while the stuck-off fault is when the gap size is stuck at 1.7 nm (~4.4 µS). The ReRAM pairs of a defective logical location are assigned the same conductance. For example, if a stuck-off fault occurs at the logical row $j$ and logic column $i$, $G_{ji}^{+} = G_{ji}^{-} = 4.4\ \mu S$. To model realistic fabrication or operational failure scenarios, we introduce six distinct spatial fault patterns, each designed to capture different types of localized or distributed defect topologies. These patterns are illustrated in Fig. 2 and are applied to individual ReRAM arrays (layers) of the neuromorphic circuit.

1) Circular Defects: A dense, solid circular region of defective cells centered in the array. The fault region is defined by a single parameter: the radius $r \in \{0.1, 0.2, 0.3, 0.4\}$, expressed as a fraction of the array width. Therefore, $r = 0.5$ means that the diameter of the circuit is the same as the array width. Increasing the radius results in a larger proportion of the array being damaged.
2) Ring Defects: A hollow annular band where defective cells form a ring. The ring is defined by a fixed outer radius $r_{outer} = 0.5$ and variable inner radii $r_{inner} \in \{0.4, 0.36, 0.32, 0.28\}$. All expressed as a fraction of the array width.
3) Circle-Complement Defects: The inverse of circular defects. Here, only the central circular region remains functional while all surrounding cells are defective. The non-defective core is parametrized by radius $r \in \{0.4, 0.3, 0.2, 0.1\}$.
4) Row Defects: A horizontal strip of consecutive defective rows introduced symmetrically around the center of the array. The size is specified by the number of rows affected, varying from 1 to 13 depending on the layer.
5) Column Defects: Similar to row defects but oriented vertically. A contiguous set of columns is disabled, with the number of defective columns ranging from 1 to 10, depending on the layer.
6) Checkerboard Defects: A spatially alternating pattern of stuck and non-stuck ReRAM cells arranged in a grid. This pattern has a fixed structure and is not parametrized by defect size.

### *C. Data Generation*

Using our simulation framework, we generate a total of 150,948 circuit-level inference samples by varying fault type, layer, and defect size. We consider six distinct defect types—Circle, Ring, Circle-Complement, Row, Column, and Checkerboard—each injected into one of four neural network layers (Layers 0–3). Except for Checkerboard defect, the other five defect types support size variation and thus we simulate four different sizes (e.g., radii 0.1 to 0.4), leading to 16 unique configurations per defect type. Each configuration produces 1,797 samples (each corresponds to one of the images), resulting in 28,752 samples per defect type. The Checkerboard defect, which has a fixed pattern and no size variation, produces 7,188 samples across its four-layer injections. Therefore, there are 150,948 data points. For each configuration of 7,188 samples, we set aside 1,000 examples for cross-validation. The remaining 6,188 samples are split 80%–20% into training and internal testing sets, yielding 4,950 training samples per configuration. Table I summarizes the sample distribution used in training and testing for each defect type.

TABLE I
DATA DISTRIBUTION FOR EACH DEFECT TYPE ACROSS ALL LAYERS

| Defect Type | Defect* Coverage | Training Samples | Test Samples | Cross-validation Samples |
|---|---|---|---|---|
| **Circular** | ~10 - 40% | 19,800 | 4,952 | 4,000 |
| **Ring** | ~10 - 40% | 19,800 | 4,952 | 4,000 |
| **Circle-comp** | ~10 - 40% | 19,800 | 4,952 | 4,000 |
| **Row** | ~10 - 40% | 19,800 | 4,952 | 4,000 |
| **Column** | ~10 - 40% | 19,800 | 4,952 | 4,000 |
| **Checker -board** | ~50% (fixed) | 4,950 | 1,238 | 1,000 |
| **All defects** | **All sizes + layers** | **103,950** | **25,998** | **21,000** |

*$^*$Defect Coverage is defined as the fraction of array cells under a defect mask. For circular/ring masks, radii are normalized to half of the smaller array dimension; values shown use the geometric area $\pi r^2$ (or $\pi(r_o^2 - r_i^2)$). Row/column percentages reflect the fraction of rows/columns marked defective. Checkerboard covers ~50% of cells. In implementation, exact coverage is computed by counting masked cells on the discrete grid.*

## III. Preliminary Defect Effect Analysis

We first define a few key variables:

1. $N$: Number of input rows of an array, including bias
2. $V_{in,j}$: Input voltage applied to row $j$
3. $G_{ji+}, G_{ji-}$: Conductance of ReRAM pairs at row $j$, column $i$+ and $i$- of logical column $i$
4. $I_{i+}$, $I_{i-}$: Currents through the positive and negative physical ReRAM columns in logical column $i$
5. $V_{o,i}$: Output voltage for logical column $i$

### A. Output Voltage Computation

The current through each branch of a logical column is computed by summing the product of the input voltage and the ReRAM conductance across rows:

$$I_{i+} = \sum_{j=1}^{N} V_{in,j}\, G_{ji}^{+}, \qquad I_{i-} = \sum_{j=1}^{N} V_{in,j}\, G_{ji}^{-} \tag{1}$$

The final differential output voltage for column $i$ is then given by:

$$V_{o,i} = f(I_{i+} - I_{i-}) \tag{2}$$

where $f(\cdot)$ is a current-to-voltage conversion function, which is linear for analog comparators:

$$V_{o,i} = R_{load}(I_{i+} - I_{i-}) \tag{3}$$

This voltage in the last layer represents the activation for digit class $i$. The predicted digit is simply:

$$Prediction\ = \arg\max_{i}\ V_{o,i} \tag{4}$$

### B. Impact of Defects on Output Voltages

When stuck-at faults are introduced in any layer of the neuromorphic circuit, they alter the conductance values $G_{ji+}$ and $G_{ji-}$ of the affected ReRAM pairs. This distortion propagates through the circuit and may impact the output voltage associated with each class. An analysis of the impact of defects in any layer, except the last layer, is difficult. We will discuss the effect when the defect is in the last layer numerically. Let the correct class label be denoted by $y_{true}$, and the output voltage for class $i$ be $V_{o,i}$. Fig. 3 depicts an example of how circular defects in Layer 3 can alter current outputs and lead to misclassification. As the defect radius increases, damage to the ReRAM pairs weakens the differential voltage of the true class column and allows neighboring columns to dominate, shifting the prediction from 6 to 7 and then to 8. Two distinct inference error mechanisms can be observed.

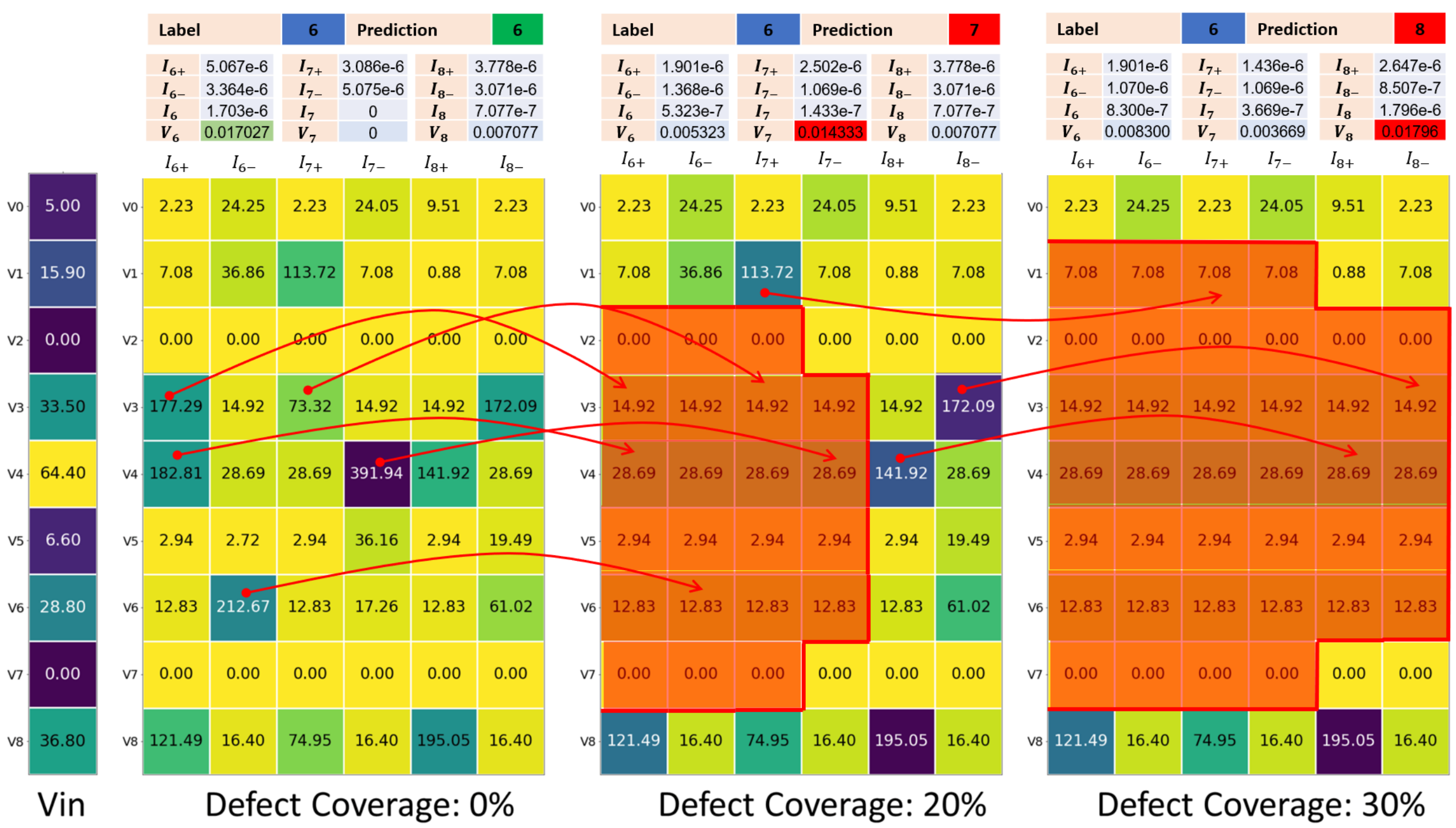

**Fig. 3:** Visualization of layer 3 slice of columns 6 – 8 current outputs under increasing circular defect severity with true label = 6. The tables above each panel report ground-truth, predicted digit, and total branch currents; the heatmaps show contributions from input voltages $V_0 - V_8$, with color representing current magnitude (normalized). Orange shaded region highlight cells impacted by the injected fault. (Left) In the defect-free case, the neuromorphic circuit correctly predicts the digit as '6' with $I_7 = 0$ after rectification of the negative differential value. (Middle) With a circular defect coverage of 20%, damage to ReRAM cells in columns $I_{6+}$, $I_{6-}$, $I_{7+}$, and $I_{7-}$ reduces the differential current in column 6 and increases that in column 7 (as the very negative difference in row V4 becomes 0 due to the defects), causing a misclassification. (Right) At a larger defect coverage of 30%, additional degradation in $I_{8+}$ and $I_{8-}$ further suppresses $V_{o,6}$, shifting the prediction to '8'.

The first occurs when the defect is located in the column corresponding to the correct class $y_{true}$. In this case, the current in that column is directly affected, changing the differential voltage for the correct class. For example, if column 6 corresponds to the true digit ($y_{true} = 6$), a fault in column 6 modifies $I_{6+}$ and $I_{6-}$. This changes the resulting voltage $V_{o,6}$, which may then fall below the voltage of another column, giving a wrong result. In this situation, the specific nature of the stuck-at fault (stuck-on or stuck-off) does not change the outcome, since both set the corresponding $G_{ji}^{+} - G_{ji}^{-}$ to 0.

The second mechanism occurs when the defect lies in a column other than the one representing the correct class, i.e., in column $j \neq y_{true}$. Here, the defect modifies the currents in that incorrect column in such a way that its output voltage increases. For instance, if $y_{true} = 6$ but the fault occurs in column 7, $G_{4,7}^{+} - G_{4,7}^{-}$ changes from 28.69 - 391.94 < 0 to 28.69 - 28.69 = 0, the defect increases $I_{7+} - I_{7-}$, thereby raising $V_{o,7}$. If the boosted voltage $V_{o,7}$ surpasses $V_{o,6}$, the circuit incorrectly predicts class 7 instead of the correct class.

### *C. Layer Dependence*

To enable quantitative comparison across different fault types, we define the defect severity by the number of defective ReRAM pairs per array. For instance, array 0 has 3250 total ReRAM pairs. A defect radius of 0.1 in a circular pattern affects approximately 12.5% of the array, or about 400 ReRAM pairs. The impact of spatial defects varies with the layer in which they are introduced, but the effect is not uniform across all patterns. As shown in Fig. 4, certain defect types, such as circle and row faults, cause the steepest degradation when injected in the output layer. For the ring and circle-complement, the degradation is more severe when the defect is in the first layer (Layer 0). These results indicate that layer sensitivity is strongly dependent on the defect geometry, with no single layer being universally most vulnerable.

## IV. Error Correction

To mitigate the inference degradation caused by stuck-at faults in ReRAM neuromorphic circuits, we introduce a lightweight post-inference neural correction model that maps faulty output voltages to corrected class labels by utilizing the distorted output voltages of the analog circuit as input features. Fig. 5 illustrates the steps of corrective inference wherein a defective neuromorphic circuit misclassifies an input digit ('8' → '9') due to stuck-at faults. And the corrective neural network restores the correct prediction by analyzing the distorted output voltage vector.

### *A. Problem Formulation*

Let the output voltage vector from the analog circuit in the last layer be denoted as:

$$V_{circuit} = [V_{o,0}, V_{o,1}, \dots, V_{o,9}]^T \in \mathbb{R}^{10} \quad (5)$$

where $V_{o,i}$ is the analog output voltage corresponding to the digit class $i$. In the absence of faults, the predicted label is:

$$\hat{y}_{original} = \arg\max_i V_{o,i} \quad (6)$$

In the presence of stuck-at-faults, this prediction becomes unreliable. We introduce a corrective neural network $F_\theta$ which learns to predict the correct label as shown:

$$\hat{y}_{corrected} = \arg\max_i F_\theta(V_{circuit})_i \quad (7)$$

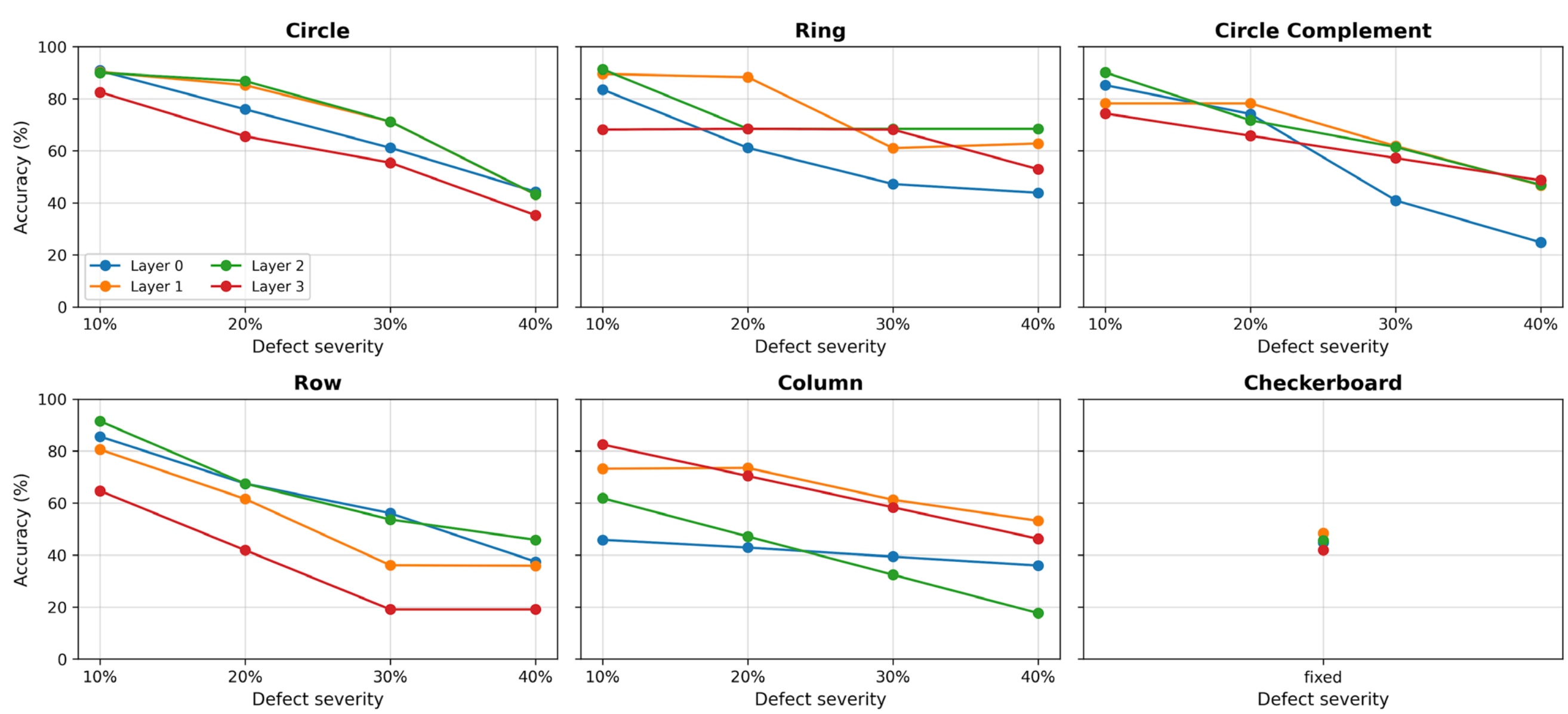

**Fig. 4:** Layer sensitivity of neuromorphic inference accuracy under different spatial defect patterns. Each panel shows classification accuracy versus defect severity for faults injected in Layers 0–3.

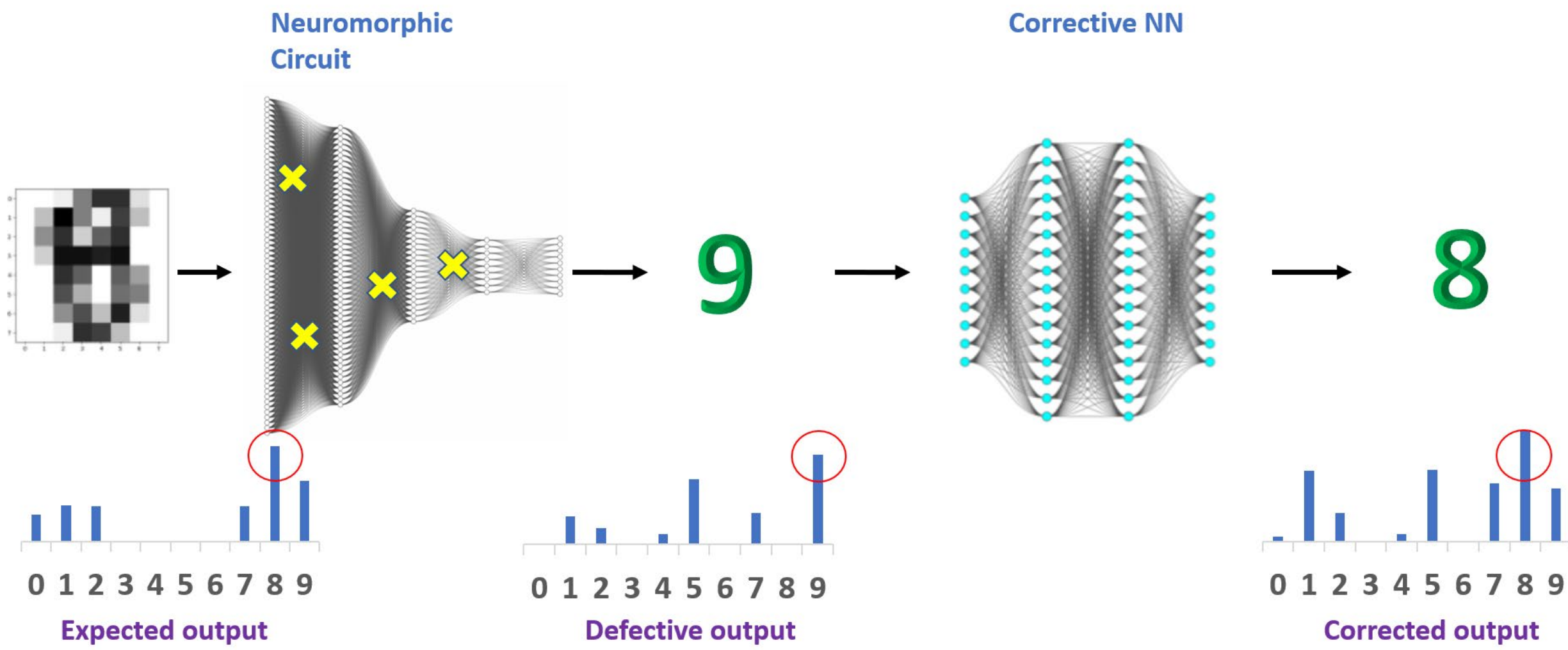

**Fig. 5:** Illustration of corrective inference. A defective neuromorphic circuit misclassifies an input digit ('8' → '9') due to stuck-at faults. A corrective neural network restores the correct prediction by analyzing the distorted output voltage vector.

where $F_\theta: \mathbb{R}^{10} \rightarrow \mathbb{R}^{10}$ is a trainable function parameterized by weights θ, mapping the faulty voltage vector to corrected class probabilities.

*B. Corrective Model Architecture & Learning Objective*

The corrective network is implemented as a multilayer perceptron (MLP) that maps the defective circuit outputs back to the correct class labels. Its input is the 10-dimensional voltage vector $V_{circuit}$ produced by the faulty neuromorphic circuit, and its output is a 10-way probability distribution over the digit classes {0, … 9}. We evaluated a broad set of corrective neural network architectures, ranging from compact single-hidden-layer models with only a few neurons (fewer than 100 parameters) to larger multi-layer perceptrons (with 3000 parameters). In the general case of a two-hidden-layer fully connected feedforward network, the input is the 10-dimensional voltage vector $V_{circuit}$, which is transformed through successive hidden layers with ReLU activation before reaching the 10-class softmax output. For hidden-layer widths, we explored configurations with 32, 16, 12, 10, 6, and 4 neurons. The network computes:

$$\mathrm{h}^{(1)} = ReLU\left(W^{(1)}V_{circuit} + \mathrm{b}^{(1)}\right) \tag{8}$$

$$\mathrm{h}^{(2)} = ReLU\left(W^{(2)}\mathrm{h}^{(2)} + \mathrm{b}^{(2)}\right) \tag{9}$$

$$\hat{y} = softmax\left(W^{(3)}\mathrm{h}^{(2)} + \mathrm{b}^{(3)}\right) \tag{10}$$

Training is performed in a supervised manner, where each sample consists of the voltage vector $V_{circuit}$ paired with the true digit label $y_i$. Although the input dimension is ten voltages, the training target is not another voltage vector but the digit identity, represented in one-hot encoded form. The model is optimized using the categorical cross-entropy loss

$$\mathcal{L}(\hat{y}, y) = -\sum_{i=0}^{9} y_i \log(\hat{y}_i) \tag{11}$$

where $y_i$ is one-hot encoded ground-truth label for class $i$, $\hat{y}_i$ is the predicted probability over the 10 digits classes from the corrective neural network. Through this formulation, training effectively learns a mapping from distorted voltage patterns to correct class predictions. The corrective network thus acts as a lightweight post-processor, compensating for systematic distortions introduced by stuck-at and spatially structured faults in the underlying ReRAM arrays.

To identify the smallest feasible corrective model for low-power edge deployment, we evaluated a range of neural network architectures as summarized in Table II. The table also summarizes their parameter counts and descriptions of complexity. For comparison, the ReRAM circuit with defects is 4528 parameters.

TABLE II
CORRECTIVE MODELS

| Model Architecture | Tunable Parameters | Complexity Category |
|---|---|---|
| MLP(100, 200) | 23,310 | Large |
| MLP (32, 64) | 3,114 | Large |
| MLP (32, 32) | 1,738 | Medium |
| MLP (16, 32) | 1,050 | Medium |
| MLP (16, 16) | 618 | Medium |
| MLP (12, 12) | 418 | Medium |
| MLP (10, 10) | 330 | Small |
| MLP (10,) | 220 | Small |
| MLP (6, 6) | 178 | Small |
| MLP (6,) | 136 | Small |
| MLP (1,) | 31 | Tiny |

## V. Evaluation Methodology

To evaluate the correction model, we measure its performance across fault types, defect sizes, and network layers under two distinct regimes. In the same-defect case, the corrective network is trained and tested on the same type and size of defect, providing a direct measure of its ability to compensate for specific fault scenarios. In the cross-defect case, the model is trained on one defect type and evaluated on another, allowing us to assess its ability to generalize to unseen fault conditions. The improvement is quantified as:

$$\Delta A = A_{corrected} - A_{faulty} \quad (12)$$

where $A_{faulty}$ is the inference accuracy of the uncorrected defective circuit, and $A_{corrected}$ is the accuracy after applying the corrective model.

### *A. Same-defect-type Correction*

In the same-defect setting, the corrective model is both trained and tested on the same defect type. For each defect type, the training and test datasets are generated by injecting faults of all four severities into each of the four layers of the neuromorphic circuit. For example, a machine is built by training circular defects with all four radii, $r \in \{0.1, 0.2, 0.3, 0.4\}$, and in different layers to correct unseen data with circular defects of any dimensions and in any layer. Table III provides the quantitative improvements achieved under the same-defect correction across all defect types.

Results show that compact two-layer MLPs (e.g., 10–10, 330 parameters) are able to substantially restore performance, achieving 30–40 percentage points (pp) of recovery for circle, ring, row, and column defects. Circle-complement and checkerboard defects remain more challenging, with improvements limited to about 15–20 pp. Therefore, small nonlinear models are sufficient to achieve strong recovery in the same-defect settings, with diminishing returns for larger architectures.

TABLE III
CORRECTION PERFORMANCE FOR THE SAME-DEFECT CASE

| Defect Type | Corrective Model | Avg. Improvement |
|---|---|---|
| **Circular, Ring, Row, Column** | Small MLP (10, 10) | +30 pp to 40 pp |
| **Circle-complement** | Small MLP (10, 10) | +15 pp to 20 pp |
| **Checkerboard** | Small MLP (10, 10) | +10 pp to 15 pp |

### *B. Cross-defect Generalization*

In the cross-defect setting, the corrective model is trained on one defect type and evaluated on a different defect type. For each experiment, the training and test datasets are generated by injecting faults of all four severities into each of the four layers of the neuromorphic circuit, ensuring that generalization is assessed across the full range of defect sizes and layer locations. Let $A_{d_1 \to d_2}^{train \to test}$ denote the accuracy when trained on defect type $d_1$ and tested on $d_2$. Each architecture in Table II was tested across both same-defect and cross-defect scenarios. Since the neuromorphic circuit itself contains approximately 4,500 trainable parameters, corrective models larger than this would defeat the purpose of lightweight post-processing and are therefore considered impractical.

Table IV summarizes these cross-defect results and Fig. 6 shows the improvements of inference accuracy for three selected $d_1$ and $d_2$ (Circle, Ring, Circle-complement) at various severities with different MLPs. The results indicate that cross-defect generalization is possible, but it is largely limited to structurally related patterns. Corrective MLPs trained on ring defects transfer reasonably well to circle defects, and vice versa, with 15–25 percentage points of accuracy recovered. Partial transfer is also observed between row and column defects, reflecting their directional similarity and yielding about 10–20 pp of improvement. In contrast, generalization between circle-complement and circle defects is poor, typically recovering less than 10 pp and in some cases reducing accuracy further due to the global nature of circle-complement faults. Checkerboard defects show almost no transferability to any other defect type (<5 pp), since their alternating pattern produces highly structured disruptions that cannot be captured by corrective models trained on unrelated patterns. Therefore, corrective models are effective when defect geometries share structural similarity but are ineffective when patterns are global or highly irregular.

TABLE IV
CORRECTION PERFORMANCE FOR CROSS-DEFECT CASE

| Defect Type | Corrective Model | Avg. Improvement |
|---|---|---|
| Ring ↔ Circle | Small MLP (10, 10) | +15pp to 25 pp |
| Row ↔ Column | Small MLP (10, 10) | +10 to 20 pp |
| Circle-comp ↔ Circle | Small MLP (10, 10) | < +10 pp |
| Checkerboard ↔ Any | Small MLP (10, 10) | < +5 pp |

Generalization effectiveness depends on how representative the training defect is in terms of output voltage distortion. The circle-complement dataset consistently underperforms in generalization tasks, especially when trained for one defect type and tested on another. This is attributed to the fact that circle-complement faults disrupt peripheral columns, introducing class-agnostic noise that reduces pattern learnability. This calls for the inclusion of hybrid datasets or augmented variants to enhance generalization.

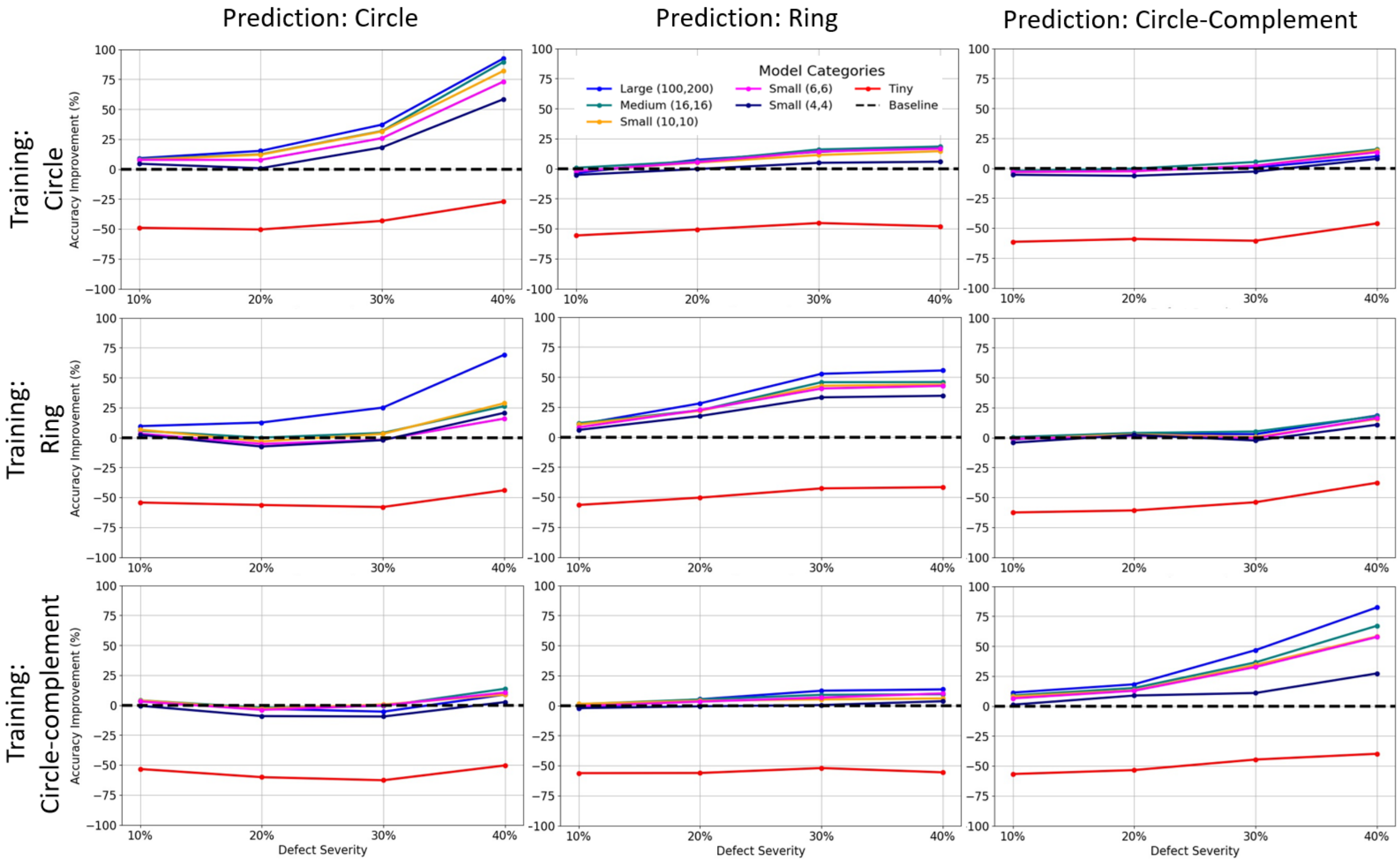

**Fig. 6:** Accuracy improvement over the neuromorphic circuit baseline for selected corrective models under different training and testing conditions. Each subplot shows the percentage improvement in classification accuracy (Y-axis) as defect severity increases (X-axis), for a range of model architectures (see legend).

## VI. Discussions

In certain situations, applying a corrective model can actually reduce performance rather than improve it. This is most evident when the model is trained on one type of defect and then tested on a structurally different pattern like a ring. Degradation can also occur when the model overfits to a particular defect geometry or layer. For example, a corrective model trained on circle-complement defects but tested on circular defects of radius 0.3 in Layer 3 reduces accuracy by about 5 percentage points compared to the uncorrected circuit (row 3 column 1 in Fig. 6). Although such cases are relatively rare, they emphasize the importance of evaluating corrective models across a broad range of defect scenarios instead of assuming reliable generalization.

We also tested an extreme case with a tiny model (31 parameters). Across all experiments, the tiny corrective model consistently degraded performance rather than improving it. As shown in Fig. 6, its accuracy fell well below the baseline defective circuit, in many cases by 20–50 percentage points. This outcome underscores that an extremely small corrective model not only lacks the capacity to represent the defect-to-output mapping, but can also amplify error, making it a clear failure case in our evaluation. These results highlight that a minimum representational capacity is necessary for effective correction (such as the two-layer MLPs with only ~300 parameters).

Remarkably, small corrective architectures achieve competitive accuracy for some of the same-defect-type corrections. For instance, the (4,4) model with only 126 parameters reached 92.4% accuracy on mild circle defects and maintained 65.6% even under severe conditions. Similarly, other compact models with fewer than 200 parameters consistently delivered strong results in same-defect correction, confirming their suitability for energy-constrained edge applications.

## VII. Conclusion

This work presented a comprehensive framework for analyzing and correcting inference errors in ReRAM-based neuromorphic circuits under spatially correlated defects. Using a DTCO simulation pipeline with SPICE-level device modeling, we systematically studied the impact of stuck-at faults across six defect patterns and four network layers. Our analysis demonstrated that both the type and location of defects significantly affect accuracy: output-layer faults directly bias specific class logits, while input-layer faults can propagate diffusely and cause equally severe degradation.

To address these challenges, we introduced compact corrective neural networks trained on faulty circuit outputs.

When trained and tested on the same defect type, shallow two-layer MLPs recovered 30–40 percentage points of lost accuracy, while circle-complement and checkerboard defects remained partially correctable. In cross-defect scenarios, generalization was effective between structurally related patterns (e.g., Circle ↔ Ring, Row ↔ Column) but poor for diffuse or alternating patterns (Circle-Complement, Checkerboard). Importantly, we also observed rare degradation cases where mismatched training and test defects reduced accuracy below the faulty baseline, underscoring the need for careful validation.

A key insight from our model size optimization is that very small models (<200 parameters) are sufficient for strong correction. While large networks yield only marginal benefits, a shallow 2-layer (10, 10) MLP consistently achieved near-optimal results. By contrast, a tiny corrective model failed entirely, confirming that minimal nonlinearity is required to capture defect-induced distortions.

Overall, this study provides both a methodology and experimental evidence that lightweight corrective models can restore substantial accuracy in defective neuromorphic circuits. These findings have direct implications for low-power edge deployment, where area and energy constraints favor compact solutions.

## IX. Acknowledgment

This material is based upon work supported by the National Science Foundation under Grant No. 2046220.

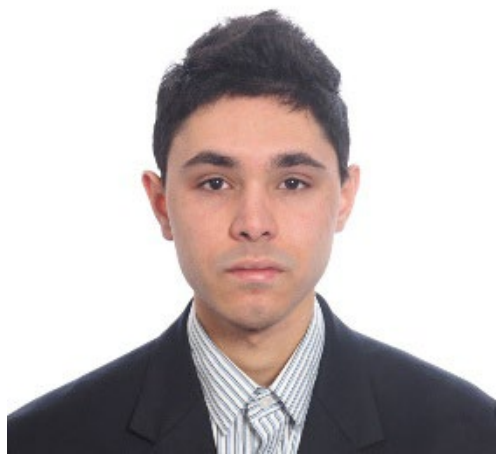

**Vedant V. Sawal** (Student Member, IEEE) received the B.S. degree in computer science, *magna cum laude*, from San José State University. He is currently pursuing the M.S. degree in computer science at SJSU, with a focus on applying AI models on a variety of use-cases and fault-tolerant machine learning systems.

He has been an intern at Sandia National Laboratories since Summer 2024, where he contributes to projects at the intersection of high-performance computing and bolide research. His research experience includes the design and simulation of analog ReRAM-based neuromorphic architectures, error correction models, and high-performance computing applications. He has co-authored papers and posters published in IEEE, The Astronomical Journal, the American Geophysical Union (AGU), and the American Astronomical Society (AAS). His current research interests include fault-tolerant computing, in-memory architectures, and edge AI systems.

Mr. Vedant Sawal is a student member of the IEEE. He is also affiliated with professional scientific organizations such as AGU and AAS, and actively participates in interdisciplinary research collaborations across computing and space sciences.

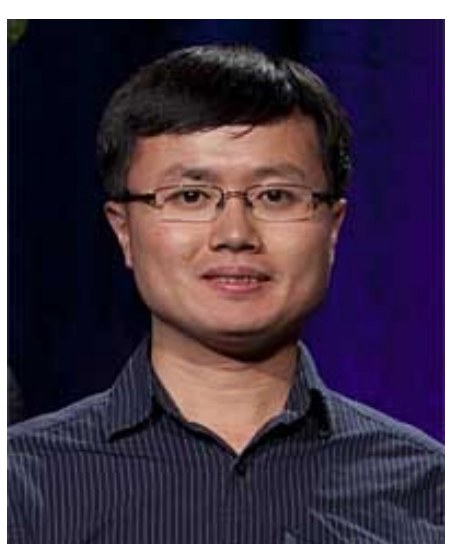

**HIU YUNG WONG** (M’12–SM’17) is a Professor at San Jose State University. He received his Ph.D. degree in Electrical Engineering and Computer Science from the University of California, Berkeley in 2006. From 2006 to 2009, he worked as a Technology Integration Engineer at Spansion. From 2009 to 2018, he was a TCAD Senior Staff Application Engineer at Synopsys.

He received the Industry Sponsored Research Award and ERFA RSCA Award in 2024, the AMDT Endowed Chair Award, the Curtis W. McGraw Research Award from ASEE Engineering Research Council in 2022, the NSF CAREER award and the Newnan Brothers Award for Faculty Excellence in 2021, and the Synopsys Excellence Award in 2010. He is the author of two books, "Introduction to Quantum Computing: From a Layperson to a Programmer in 30 Steps" and "Quantum Computing Architecture and Hardware for Engineers: Step by Step". He is one of the founding faculty members of the Master of Science in Quantum Technology at San Jose State University.

His research interests include the application of machine learning in simulation and manufacturing, cryogenic electronics, quantum computing, and wide bandgap device simulations. His works have produced 2 books, 1 book chapter, more than 130 papers, and 10 patents.